\documentclass[letterpaper, 10pt, conference]{ieeeconf}  

\IEEEoverridecommandlockouts                              

\usepackage{amsmath}
\usepackage{amssymb}	
\usepackage{mathtools}  

\usepackage{algorithm}
\usepackage{algcompatible}
\usepackage{algorithmicx}
\usepackage{algpseudocode}

\usepackage{graphicx}
\usepackage{epstopdf}
\graphicspath{diagrams}
\usepackage{caption}
\usepackage{subcaption}
\usepackage{float}
\usepackage{multirow}
\usepackage{tikz}	  

\title{\LARGE \bf
Goal-based Self-Adaptive Generative Adversarial Imitation Learning (Goal-SAGAIL) for Multi-goal Robotic Manipulation Tasks
}

\author{Yingyi Kuang$^{1}$, Luis J. Manso$^{1}$ and George Vogiatzis$^{2}$
\thanks{$^{1}$Y Kuang and L J Manso are with the College of Engineering and Physical Sciences, 
        Aston University, Birmingham, UK. Emails:
        {kuangy3@aston.ac.uk; l.manso@aston.ac.uk.}}
\thanks{$^{2}$G Vogiatzis is with the Department of Computer Science Department, Loughborough University, UK. Email:
         {g.vogiatzis@lboro.ac.uk}}%
}

\begin{document}

\maketitle
\thispagestyle{empty}
\pagestyle{empty}

\begin{abstract}

Reinforcement learning for multi-goal robot manipulation tasks poses significant challenges due to the diversity and complexity of the goal space. 
Techniques such as Hindsight Experience Replay (HER) have been introduced to improve learning efficiency for such tasks. More recently, researchers have combined HER with advanced imitation learning methods such as Generative Adversarial Imitation Learning (GAIL) to integrate demonstration data and accelerate training speed.   
However, demonstration data often fails to provide enough coverage for the goal space, especially when acquired from human teleoperation.
This biases the learning-from-demonstration process toward mastering easier sub-tasks instead of tackling the more challenging ones.
In this work, we present Goal-based Self-Adaptive Generative Adversarial Imitation Learning (Goal-SAGAIL), a novel framework specifically designed for multi-goal robot manipulation tasks. By integrating self-adaptive learning principles with goal-conditioned GAIL, our approach enhances imitation learning efficiency, even when limited, suboptimal demonstrations are available. 
Experimental results validate that our method significantly improves learning efficiency across various multi-goal manipulation scenarios---including complex in-hand manipulation tasks---using suboptimal demonstrations provided by both simulation and human experts.

\end{abstract}

\section{INTRODUCTION}

In recent decades, reinforcement learning (RL) has progressed considerably in various robot manipulation scenarios.
Although several studies have successfully applied RL to practical robot manipulation tasks~\cite{kalashnikov2018scalable,andrychowicz2018learning}, multi-goal RL in continuous task domains remains a significant challenge.
In multi-goal RL~\cite{plappert2018multi}, agents are required to tackle a range of related tasks, each characterised by its own distinct reward function.
Furthermore, in multi-goal RL, the reward is frequently only provided upon task success, resulting in sparse and delayed feedback.
The complexity and diversity of the goal space pose yet another challenge that multi-goal agents have to overcome.
An agent needs to generalise across a wide range of goals---each potentially having very different dynamics and success criteria---while receiving infrequent feedback (rewards) only when a goal is achieved. This poses challenges to both effective exploration and sample-efficient learning.
Consequently, these agents tend to encounter frequent failures and predominantly negative sparse reinforcement, especially during early training stages. 

\par
For continuous multi-goal RL challenges, methods such as \emph{Hindsight Experience Replay} (HER)~\cite{andrychowicz2017hindsight} have been introduced to enhance data efficiency. 
HER enables the agent to learn from its failures by retrospectively reassigning \textit{``achieved''} goals from past trajectories as targets, thereby recalculating rewards and substantially increasing the proportion of positive feedback. This mechanism accelerates learning, particularly in environments where rewards are sparse and delayed.

Imitation Learning (IL), also known as Learning from Demonstration (LfD), provides an alternative strategy to address sampling inefficiency. 
\emph{Generative Adversarial Imitation Learning} (GAIL)~\cite{ho2016generative} combines IL with the principles of Generative Adversarial Networks (GANs). GAIL employs a discriminator network to evaluate the differences between data generated by the training agent and the expert demonstrations.
The feedback from the discriminator is then used as a reward signal to train the agent (acting as the generator) with standard reinforcement learning techniques, ultimately guiding it to generate outputs that closely resemble the expert's behaviour. 
Later developments introduced \emph{Goal-conditioned Generative Adversarial Imitation Learning} (Goal-GAIL)~\cite{ding2019goal}, which extends GAIL to tasks conditioned on multiple goals. By integrating HER, Goal-GAIL employs HER's experience relabeling to broaden the coverage of the goal space during both the discriminator training and the conventional RL generator training processes. This integration results in a notably faster convergence in numerous multi-goal robotic learning scenarios compared to the original GAIL approach~\cite{cai2020goalconditioned}.


The original GAIL algorithm is highly dependent on both the quality and quantity of available demonstrations. 
However, in practice, obtaining a sufficient number of high-quality demonstrations is challenging, as the collected data is often limited and suboptimal, which can constrain the agent's performance to merely mimic the examples provided. 
To overcome this challenge, several approaches have been proposed~\cite{guo2018generative,shen2022learning,peng2024reinforcement} that integrate self-imitation learning into the GAIL framework.
Our method takes inspiration from one such approach, \emph{Self-adaptive Generative Adversarial Imitation Learning} (SAIL)~\cite{zhu2022self}, which incorporates high-quality trajectories generated by the actor during training as additional viable demonstrations.
SAIL continuously evaluates self-generated trajectories against expert demonstrations, identifying those that outperform the suboptimal expert examples. These better-performing trajectories then replace the earlier demonstrations in subsequent training rounds.
However, SAIL’s original design does not readily extend to multi-goal learning problems. In these scenarios, the diverse difficulties of different goals make it challenging to compare episode rewards directly, as such comparisons must be conditioned on the specific goal. As a result, the simple approach of selecting high-quality trajectories is insufficient in multi-goal settings.

In this paper, we introduce \emph{Goal-based Self-Adaptive Generative Adversarial Imitation Learning}, or Goal-SAGAIL. 
Drawing inspiration from SAIL, our method extends self-adaptive learning principles to goal-conditioned GAIL specifically for multi-goal tasks, thereby improving learning efficiency even when limited suboptimal demonstrations are available.
The key innovation of our approach is a goal-based mechanism that selects high-quality, self-generated trajectories relative to the provided demonstrations in multi-goal robotic manipulation tasks. 
These superior trajectories are then integrated into the demonstration dataset, facilitating a smooth and effective self-adaptive transition throughout the learning process. 

We evaluated Goal-SAGAIL across various multi-goal robotic manipulation scenarios, including several challenging in-hand manipulation tasks from the Gymnasium-Robotics simulation environment~\cite{gymnasium_robotics2023github}.  
Additionally, we applied our method to use data acquired through teleoperation with a human demonstrator. 
Specifically, we collected human-teleoperated demonstrations using Leap Motion for a block rotation task with a Shadow Hand, and used these data to conduct LfD. 
The comparisons, performed across several imitation learning strategies on these LfD tasks, indicate that Goal-SAGAIL outperforms the original Goal-GAIL when dealing with suboptimal demonstrations.

\section{METHODOLOGY}
\label{methodology}

\subsection{\textbf{Preliminaries}}

\subsubsection{\textbf{Multi-goal learning}}

Consider a learning problem framed as a Markov decision process (MDP) denoted by $(\mathcal{S}, \mathcal{A}, R, P, p(s_0))$, where $\mathcal{S}$ is the state space and $\mathcal{A}$ is the action space.
Each goal $g$ is drawn from a goal space $\mathcal{G}$, and can be obtained from a state $s$ using the mapping $g = f(s)$, the goal space is a subset of the state space ($\mathcal{G} \subset \mathcal{S}$). Every episode begins with an initial desired goal $g_d \in \mathcal{G}$, which is set along with the initial state $s_0$ and remains unchanged throughout the episode. The reward function depends not only on states and actions but is also conditioned on the goal. The objective of multi-goal learning is to develop a universal goal-conditioned policy, $\tau(s, g_d)$, that maximises the cumulative reward as defined by the goal-specific reward function $R_g$ over the entire goal space $\mathcal{G}$.

\subsubsection{\textbf{Hindsight Experience Replay (HER)}}
In a multi-goal RL setup, an achieved goal $g'$ is extracted from the states $s$ at each time step of an episode. 
The reward of each time-step $t$ is determined by comparing the achieved goal at the next time step, $g'_{t+1}$, with the desired goal $g_d$, Specifically, the reward function is defined as $r_{g_d}(t) = -1{|g'_{t+1}-g_d| < \epsilon}$, where $\epsilon$ is a fixed tolerance. 
In this formulation, the reward is -1 if the achieved goal is not within the tolerance of the desired goal (unsuccessful), and 0 if it is (successful). 
As an off-policy algorithm, HER stores experiences in a replay buffer as tuples $(s_t, a_t, g'_t, g_d)$, where the desired goal $g_d$ remains unchanged throughout the episode. 
During the replay stage, some experiences are modified to use the achieved goals from the future time steps as substitute goals $g_s$ to replace the original desired goals $g_d$.
an achieved goal from the future time-step $g'_{future}$ of the same episode is used as substitute goal $g_s$ to replace the original desired goal $g_d$. 
The reward of these experiences will be calculated based on the substitute goals.
$r_{g_s}(t) = -[f(g'_{t+1}) = 0], f(g') = |g'-g'_{future}| < \epsilon$. 
Using this approach, the RL algorithm can adapt to and achieve a goal even if it differs from the one originally intended.


\subsubsection{\textbf{Generative Adversarial Imitation Learning (GAIL)}}


In GAIL, a discriminator $D(s,a)$ is trained to fit the expert state-action distribution and distinguish the expert transitions, $(s,a)\sim\tau_{expert}$, from the agent transitions, $(s,a)\sim\tau$. $D$ is trained to minimise:
\begin{equation}
\centering
\label{eq_GAIL_discriminator_loss}
\begin{aligned}
    L_{D(s,a)} &= E(s,a)\sim\tau[logD(s,a)] \\
    &+ E(s,a)\sim\tau_{expert}[log(1 - D(s,a))]
\end{aligned}    
\end{equation}

The agent policy $\tau$ is treated as a generative model $G$ to be trained to confuse the discriminator so that, eventually, the agent policy is good enough that the discriminator cannot differentiate it from the expert.
The link between the discriminator model $D$ and the generative model $G$ (the agent $\tau$) is to use the output of $logD(s,a)$ as the reward to train the agent policy to maximise $E(s,a)\sim\tau [logD(s, a)]$.



\subsubsection{\textbf{Goal-conditioned GAIL}}


Goal-GAIL integrates GAIL with the off-policy method Deep Deterministic Policy Gradient (DDPG) and HER. The integration of DDPG and HER has already shown promising learning capabilities in a variety of multi-goal robotic tasks. HER is implemented during the experience replay phase to ensure a robust positive reinforcement signal.
Moreover, Goal-GAIL extends this approach by incorporating HER-based experience relabelling into demonstration sampling, effectively enlarging the expert demonstration dataset. 
On the GAIL side, the loss function to train the discriminator $D$ has the same structure as GAIL. However, it is also goal-conditioned. The discriminator $D(s,a)$ is trained to minimize:
\begin{equation}
\centering
\label{eq_goalGAIL_discriminator_loss}
\begin{aligned}
    L_{D(s,g,a)} &= E(s,g,a)\sim\tau[logD(s,g,a)] \\ 
    &+ E(s,g,a)\sim\tau_{expert}[log(1 - D(s,g,a))]
\end{aligned}
\end{equation}

Rather than using the discriminator's output $D$ directly as the reward for RL, Goal-GAIL integrates it with the normal binary reward received from the agent to balance self-learning by the agent and learning from demonstrations. 

\subsection{\textbf{Goal-based self-generated trajectory selection}}
In Goal-SAGAIL, we exclusively incorporate new expert demonstrations from successful self-generated trajectories. When a new successful trajectory is obtained, Goal-SAGAIL locates a trajectory in the expert dataset that shares a comparable difficulty level, so that it can directly compare their episode rewards.  
In order to represent a trajectory's difficulty, we utilise the concept of a goal-pair introduced in our previous work~\cite{kuang2020goal}. Specifically, a trajectory's goal-pair is defined as $gp = [g_{init},g_o]$, where $g_{init}$ is the initial achieved goal and $g_d$ is the original target goal.
By comparing the goal-pairs of two trajectories, we can effectively assess whether they exhibit similar levels of difficulty.

For each newly collected success trajectory $\tau^i$, it is associated with a goal-pair $gp^i = [g^i_{init}, g^i_d]$. The combined goal-pair distance between $\tau^i$ and an expert trajectory $\tau^e$, where the expert trajectory has its own goal-pair $gp^e = [g^e_{init}, g^e_d]$, is calculated as:
\begin{equation}
\label{eq:Goal_GAIL_combined_distance}
    d_{comb}(\tau_i, \tau_e) = d(g^i_{init}, g^e_{init}) + d(g^i_d, g^e_d)
\end{equation}

For each successful self-generated trajectory, we calculate the combined goal-pair distance to every expert trajectory in the expert buffer. The expert trajectory with the minimum combined goal pair distance, denoted as $\tau_{e(min)}$, is deemed to have the most similar difficulty level to the successful self-generated trajectory $\tau_i$:
\begin{equation}
\label{eq:Goal_GAIL_most_similar_demo}
    \tau_{e(min)} = \tau_e \sim min[d_{comb}(\tau_i, \tau_e)~ |~ \tau_e \in R_E]
\end{equation}

The episode cumulative returns of $\tau_i$ and $\tau_{e(min)}$ are then compared. If the episode cumulative return of $\tau_i$ exceeds that of $\tau_{e(min)}$, it is deemed a higher-quality trajectory compared with the expert trajectory $\tau_{e(min)}$. 

In many situations, the goal-pair distribution within the expert demonstrations is not uniformly distributed, often resulting in clustered demonstrations, with a relatively easy level of difficulty. This scenario is particularly common with suboptimal experts who can only offer success trajectories of lower difficulty levels. Consequently, the expert trajectory $\tau_{e(min)}$ identified as most similar to $\tau^i$, might still have a high combined goal-pair distance, rendering it substantially different from $\tau^i$. Under these circumstances, comparing the episode returns of these two trajectories becomes less meaningful. 

To address this issue, we introduce a threshold, $C_{comb}$ during the search for the minimum combined goal-pair distance. If the smallest combined goal-pair distance found in the expert buffer exceeds this threshold, $d_{comb}(\tau_i, \tau_{e(min)}) > C_{comb}$, it indicates that the trajectory $\tau^i$ significantly diverges from any existing demonstrations and can therefore be directly considered as a high-quality self-generated trajectory.

\subsection{\textbf{Algorithm of Goal-SAGAIL}}

\begin{algorithm}[t]
\caption{Goal-based Self-Adaptive Generative Adversarial Imitation Learning (Goal-SAGAIL)}
\label{alg:algorithm1}
\begin{algorithmic}[1]
    \Require
    \Statex $\mathbb{A}$: Off-policy algorithm (e.g., DDPG)
    \Statex $\mathbb{D}$: Discriminator network (e.g., MLP)
    \Statex A goal-based reward function 
            $r: \mathcal{S} \times \mathcal{G} \times \mathcal{A} \to \mathbb{R}$ 
    \Statex A self-collected replay buffer $\mathbb{R}_B$ and an expert replay buffer $\mathbb{R}_E$
    \State Initialize actor and critic networks for $\mathbb{A}$, and the discriminator network $\mathbb{D}$
    \State Initialize $\mathbb{R}_B$; Initialize $\mathbb{R}_E$ with expert trajectories

    \State \Comment{Step 1: Data collection and storage}
    \For{$episode = 1$ to $M$} 
        \State Collect trajectory $\tau$ using the current policy of $\mathbb{A}$
        \State Find a similar-difficulty expert trajectory $\tau_{e(\min)}$ 
               using Eqs.~\eqref{eq:Goal_GAIL_combined_distance}, 
               \eqref{eq:Goal_GAIL_most_similar_demo}
        \If{$d_{comb}(\tau, \tau_{e(min)}) \leq C_{d_{comb}}$}
            \If{episode return of $\tau$ is higher than $\tau_{e(min)}$}
                \State $\tau \rightarrow \mathbb{R_E}$             
            \Else
                \State $\tau \rightarrow \mathbb{R_B}$
            \EndIf
        \Else
            \State Directly $\tau \rightarrow \mathbb{R_E}$
        \EndIf
    \EndFor

    \State \Comment{Step 2: Train discriminator}
    \For{$batch = 1$ to $N_d$} 
        \State Sample expert transitions from $\mathbb{R}_E$
        \State Sample non-expert transitions from $\mathbb{R}_B$
        \State Train $\mathbb{D}$ using Eq.~\eqref{eq_goalGAIL_discriminator_loss}
    \EndFor

    \State \Comment{Step 3: Policy update with HER}
    \For{$batch = 1$ to $N$} 
        \State Sample mini-batch $\mathbb{B}$ from $\mathbb{R}_B \cup \mathbb{R}_E$
        \State Apply HER on transitions in $\mathbb{B}$, recalculate rewards
        \State Add GAIL reward for all transitions using Eq.~\eqref{eq:Goal_GAIL_reward_function}
        \State Perform one step of optimization for $\mathbb{A}$ using $\mathbb{B}$
    \EndFor

\end{algorithmic}
\end{algorithm}

Goal-SAGAIL builds on the Goal-GAIL learning framework, with its pseudo-code detailed in Algorithm~\ref{alg:algorithm1}. Compared with Goal-GAIL, Goal-SAGAIL introduces additional steps during the data collection and storage phases (lines 6-15) to detect and preserve higher-quality trajectories as expert demonstrations.

Goal-SAGAIL maintains two replay buffers: $\mathbb{R}_E$ for expert demonstration trajectories and $\mathbb{R}_B$ for self-generated trajectories. 
The expert buffer $\mathbb{R}_E$ has been augmented with high-quality self-generated trajectories to improve its alignment with self-adaptive learning.
To avoid an overload of outdated samples, $\mathbb{R}_E$ is managed using a FIFO approach and is limited to a maximum buffer size. This is to ensure the buffer remains up-to-date by replacing older, potentially suboptimal demonstrations.
$R_B$ stores the rest of the self-generated trajectories that are considered non-expert.
During the experience replay stage for policy update, samples are uniformly drawn from a combined pool of $\mathbb{R}_E$ and $\mathbb{R}_B$, with HER applied. 

The discriminator $\mathbb{D}$ is trained in the same manner as in Goal-GAIL, with the corresponding loss function defined in Equation~\eqref{eq_goalGAIL_discriminator_loss}.
For training the discriminator, experiences are sampled using HER for both expert and self-generated experiences. The output of the discriminator is then integrated into the RL reward function. Although there are several options for using the discriminator’s output, such as $D(s,g,a)$, $log(sigmoid(D(s,g,a)))$, or $sigmoid(D(s,g,a))$. Our experiments employ $D(s,g,a)$ as the GAIL reward. The combined reward used for learning is as follows:
\begin{equation}
\centering
\label{eq:Goal_GAIL_reward_function}
    r_{combined} = (1 - \delta_{GAIL}) \cdot r_{env} + \delta_{GAIL} \cdot D(s,g,a)
\end{equation}

\noindent where $r_{env}$ is the normal RL reward received from the environment feedback. The GAIL weight $\delta_{GAIL}$ controls how much the GAIL reward affects the learning.

\section{EXPERIMENTS}

\subsection{LfD with demonstrations produced by RL-trained agent}

\subsubsection{\textbf{Environments}}
\begin{figure}[h]

    \begin{subfigure}[t]{0.45\linewidth}
      \includegraphics[width=\linewidth]{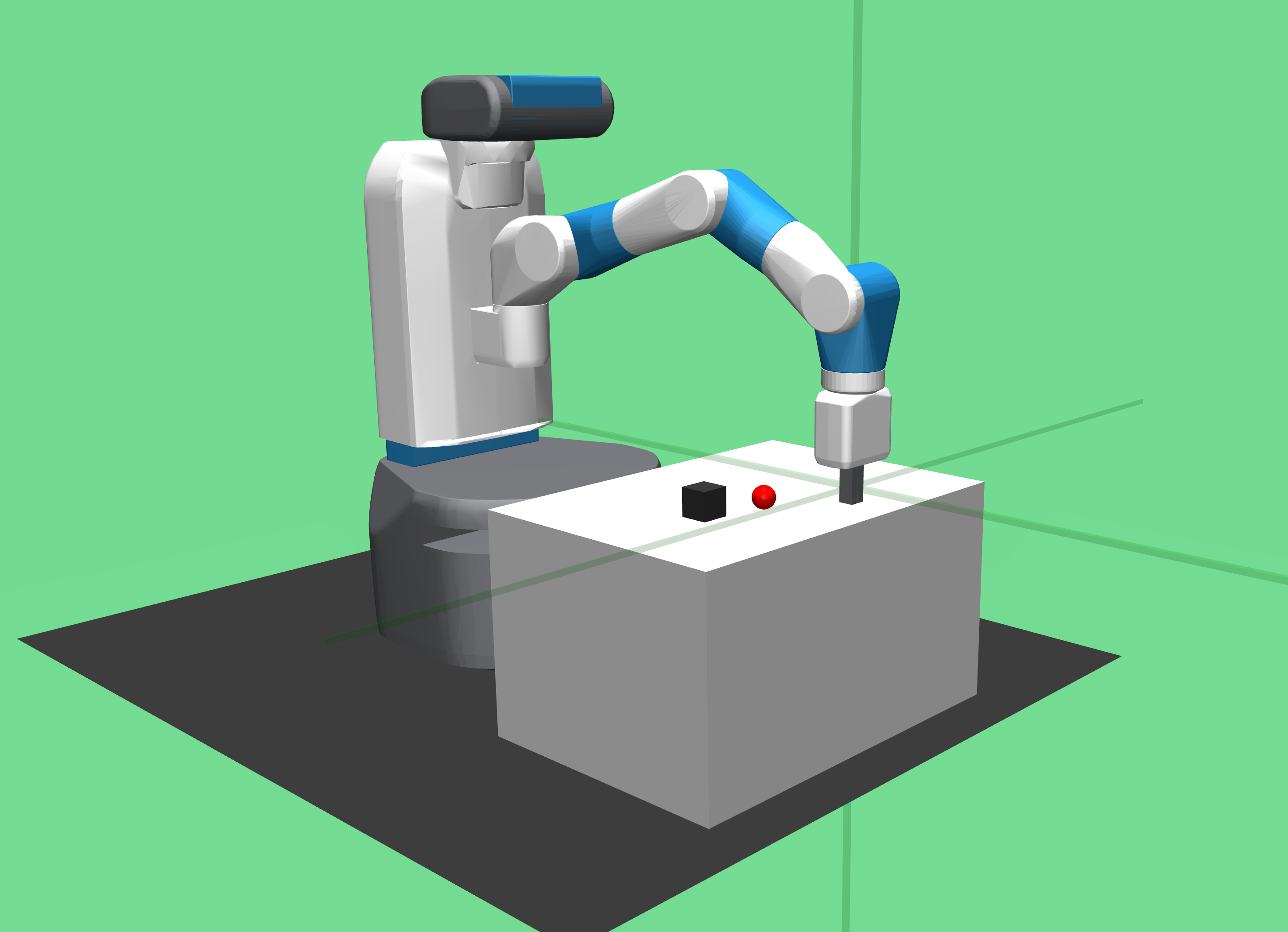} 
      \caption{FetchPush}
      \label{fig:gym-a}
    \end{subfigure}
    ~
    \begin{subfigure}[t]{0.45\linewidth}
      \includegraphics[width=\linewidth]{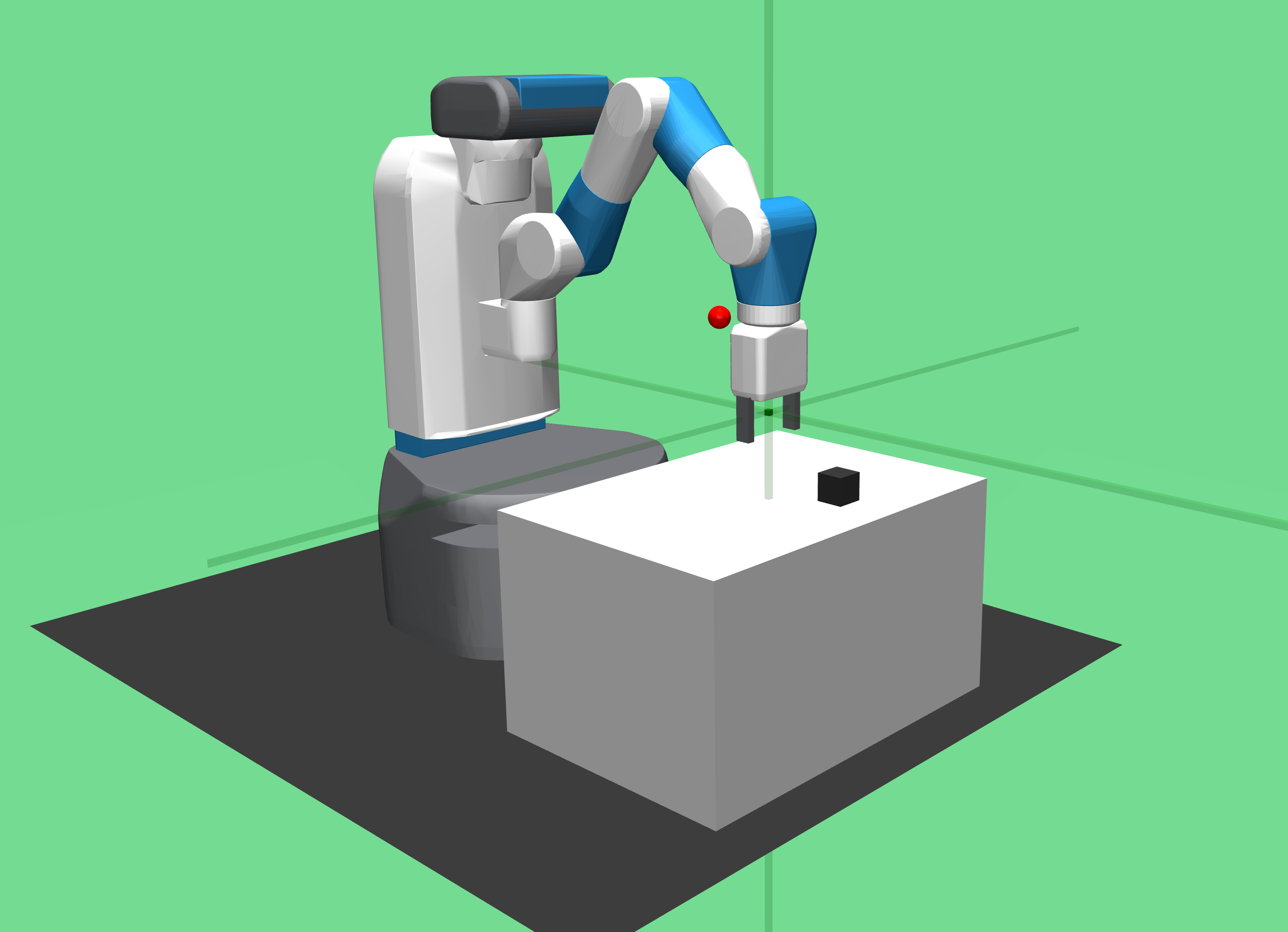}  
      \caption{FetchPickPlace}
      \label{fig:gym-b}
    \end{subfigure}
    
    \begin{subfigure}[t]{0.45\linewidth}
      \includegraphics[width=\linewidth]{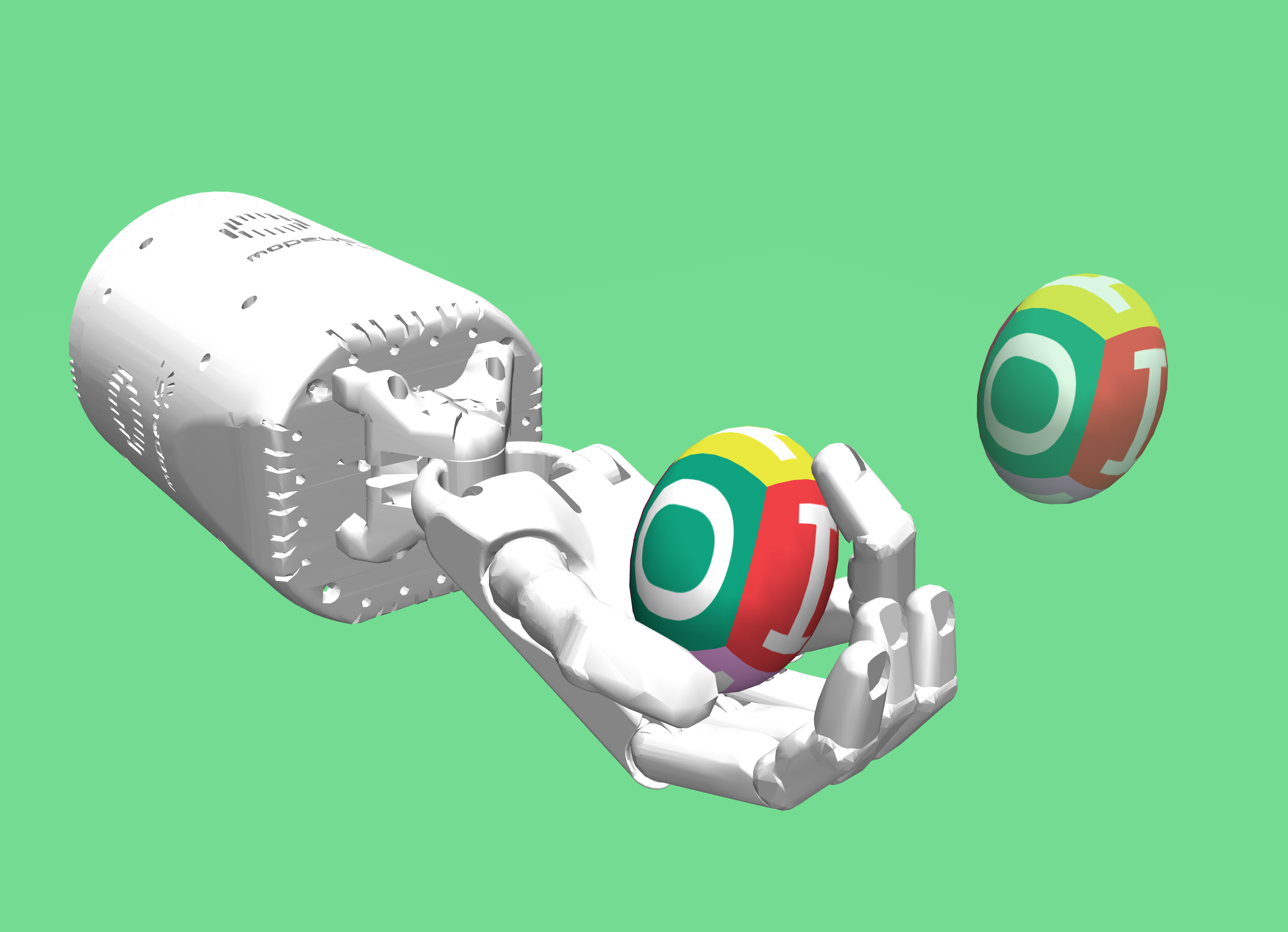}  
      \caption{HandEggRotate}
      \label{fig:gym-c}
    \end{subfigure}
    ~
    \begin{subfigure}[t]{0.45\linewidth}
      \includegraphics[width=\linewidth]{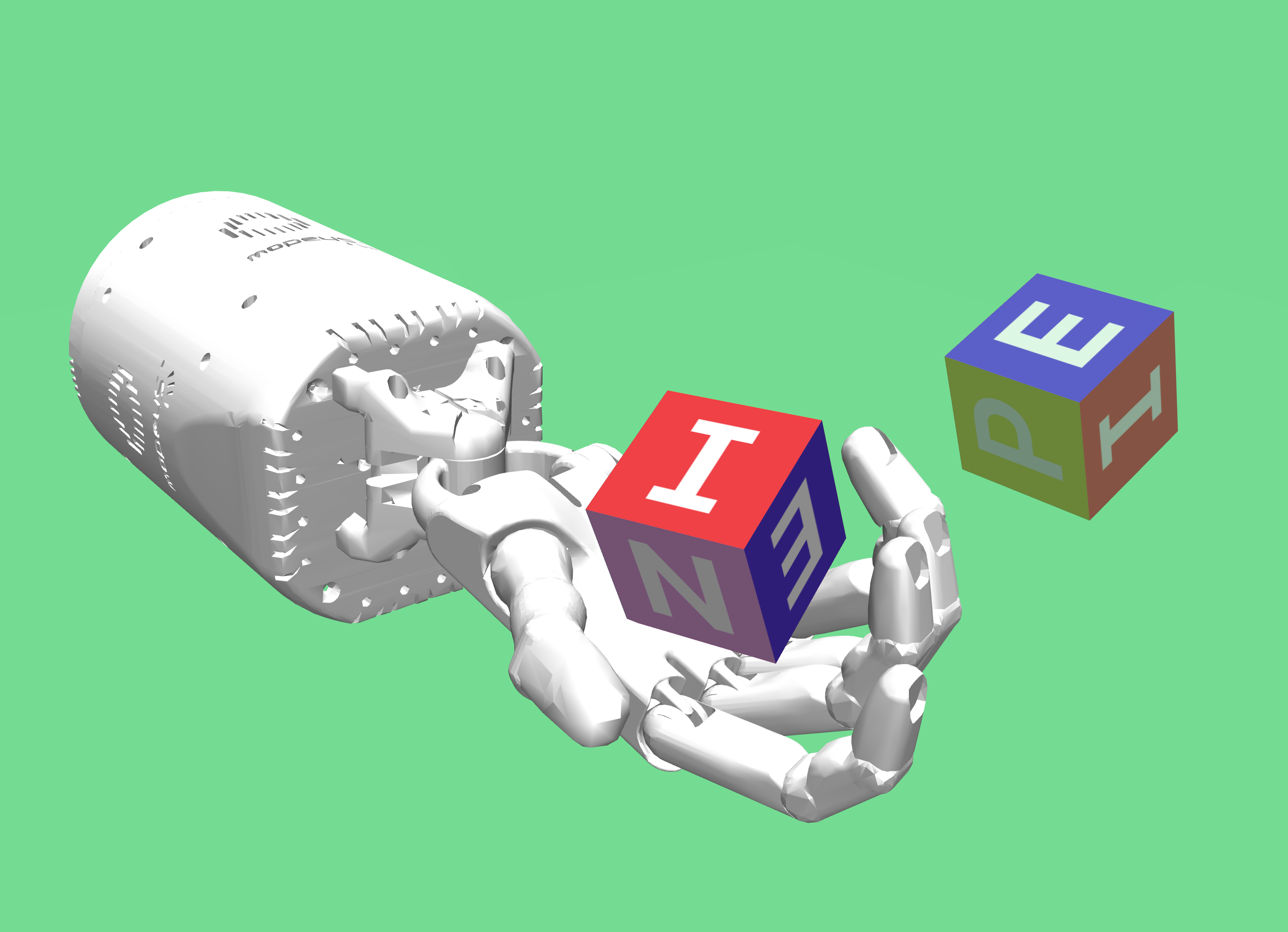}  
      \caption{HandBlockRotate}
      \label{fig:gym-d}
    \end{subfigure}
 
  \caption{Gymnasium-Robotics environments~\cite{gymnasium_robotics2023github}}
  \label{fig:gymnasium_robotics_multigoal_envs}
\end{figure}

We first tested Goal-SAGAIL on various multi-goal robotic manipulation tasks in Gymnasium-Robotics environments provided by~\cite{gymnasium_robotics2023github}, see Figure~\ref{fig:gymnasium_robotics_multigoal_envs}:

\begin{itemize}
    \item \textbf{FetchPush-v1}: a fetch robot has to move a block on a table to a designated target position (Figure~\ref{fig:gym-a}).
    \item \textbf{FetchPickAndPlace-v1}: a fetch robot has to pick up a block and move it to a specified 3D position (Figure~\ref{fig:gym-b}).
    \item \textbf{HandManipulateEggRotate-v1}: a Shadow Hand rotates an egg-shaped object within the hand to achieve a target orientation (Figure~\ref{fig:gym-c}).
    \item \textbf{HandManipulateBlockRotateXYZ-v1}: a Shadow Hand rotates a block-shaped object within the hand to achieve a target orientation (Figure~\ref{fig:gym-d}).
\end{itemize}

\subsubsection{\textbf{Hyperparameters for learning}}
We employ a suboptimal expert agent, trained via RL, to generate demonstration trajectories. These expert demonstrations are produced by using random goals and retaining only the successful trajectories. Specifically, we use 10 demonstration trajectories to initialize training for the FetchPush and FetchPickAndPlace tasks, and 400 trajectories for the HandEggRotation and HandBlockRotation tasks.

For the RL training component, each epoch consists of 50 training cycles. In every cycle, we first collect data from 40 episodes and then train the policy networks over 40 batches, with each mini-batch containing 5120 samples.  
After completing each epoch, we assess performance through evaluations over 100 episodes.

In both Goal-GAIL and Goal-SAGAIL, we use a GAIL weight $\delta_{GAIL}$ of 0.5, which is then annealed during training. The discriminator undergoes training for 40 batches per cycle, with a mini-batch size of 512.
Both the policy and discriminator networks are implemented as multilayer perceptrons (MLPs) with four layers, each having 256 nodes.

\begin{figure}
    \centering
    \begin{subfigure}[t]{0.48\linewidth}
      \includegraphics[width=\linewidth]{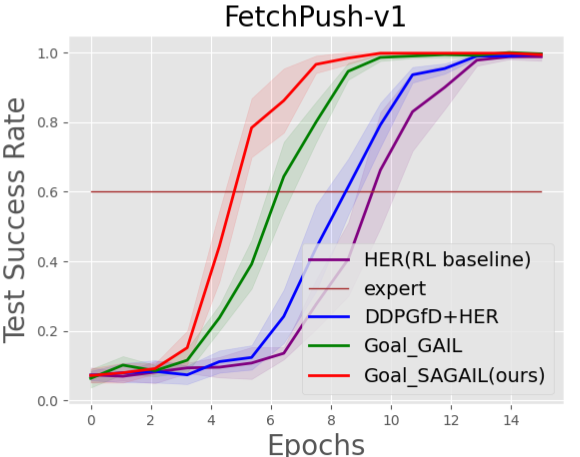}  
      \label{fig:Results_gymrobotics_subfirst}
    \end{subfigure} 
    ~
    \begin{subfigure}[t]{0.455\linewidth}
      \includegraphics[width=\linewidth]{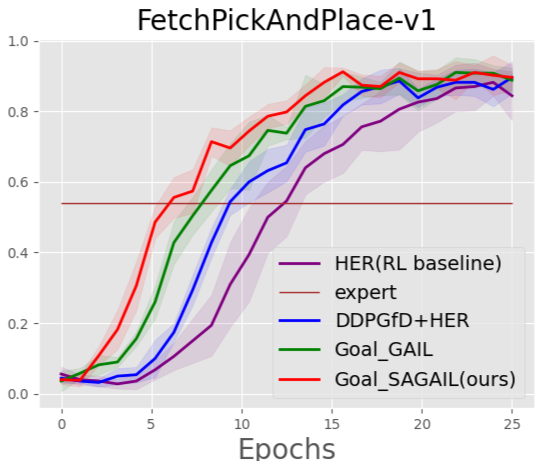}  
      \label{fig:Results_gymrobotics_subsecond}
    \end{subfigure}

    \begin{subfigure}[t]{0.48\linewidth}
      \includegraphics[width=\linewidth]{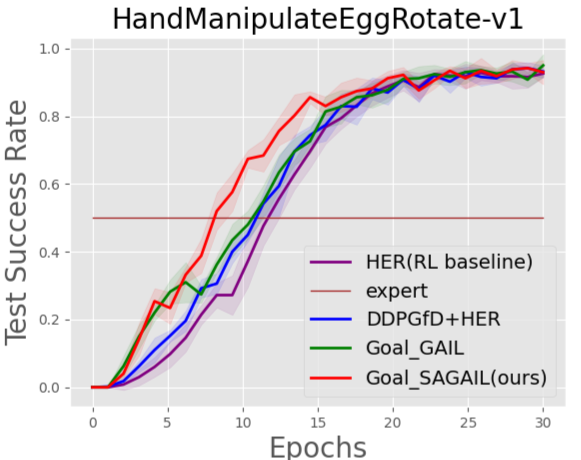}  
      \label{fig:Results_gymrobotics_subthird}
    \end{subfigure}
    ~
    \begin{subfigure}[t]{0.455\linewidth}
      \includegraphics[width=\linewidth]{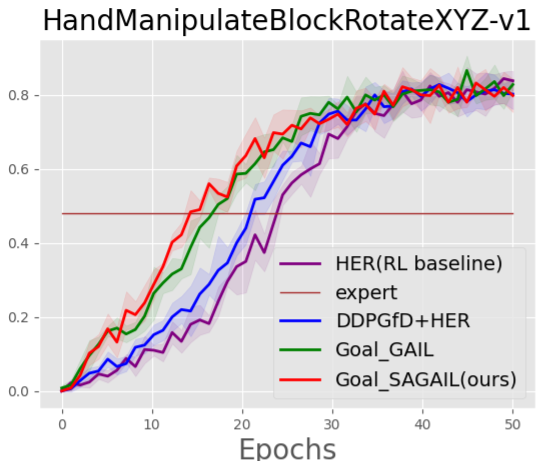}  
      \label{fig:Results_gymrobotics_subfourth}
    \end{subfigure}
  \caption[The learning curve for different LfD methods with demonstrations provided by semi-trained RL agents.]{The learning curve for different LfD methods with demonstrations provided by semi-trained RL agents. The purple, blue, green and red lines represent the learning curve for the RL baseline DDPG+HER, DDPGfD+HER, Goal-GAIL, and Goal-SAGAIL, respectively. The brown line represents the average success rate for the expert agent that is used to produce the demonstrations.}
   \label{fig:Results_gymrobotics}
\end{figure}

For Goal-SAGAIL specifically, the expert buffer $R_E$ is capped at 20 times the size of the initial demonstration data across all tasks.
The calculation of the combined goal-pair distance incorporates the task-specific goal distance $d(g_a, g_b)$: FetchPush and FetchPickPlace tasks, which involve object position manipulation, use the Euclidean distance; whereas HandEggRotate and HandBlockRotate tasks, involving in-hand object rotation, utilise the quaternion rotation difference.
Additionally, the threshold $C_{comb}$ for checking the combined goal-pair distance is set based on the task type: 0.02 for Fetch tasks and 0.25 for HandRotation tasks. 
Each experiment is repeated using five different random seeds per environment or LfD method.

\subsubsection{\textbf{Results and discussions}}

We evaluate our proposed Goal-SAGAIL method against HER (as the RL baseline), and the LfD methods DDPGfD+HER and Goal-GAIL in two Fetch robot manipulation tasks and two Shadow Hand robot in-hand manipulation tasks.  
Although the RL baseline already achieves near-optimal performance within the given time frame in all four environments, our primary focus is on how different LfD methods enhance convergence speed.  
The learning curves comparing these methods are shown in Figure~\ref{fig:Results_gymrobotics}.
The results indicate that Goal-SAGAIL considerably outperforms the RL baseline as well as both LfD methods (DDPGfD+HER and Goal-GAIL) across all environments.
In the HandBlockRotation task, while the convergence speed of Goal-SAGAIL is similar to that of Goal-GAIL, it still performs slightly better during the middle learning stage (epochs 10–20). 

Upon reviewing the demonstration dataset, we observed that while the suboptimal demonstrations do not comprehensively cover the entire goal space, they are particularly effective at achieving simpler goals with optimal movements.
Consequently, these trajectories offer ample guidance, enabling all LfD methods to rapidly master simpler tasks. The primary advantage of Goal-SAGAIL, however, lies in its ability to select new expert trajectories for more challenging and previously unseen goals.
However, in practical applications where an RL agent is not available and demonstrations are typically collected through kinesthetic teaching or human teleoperation, the quality of these suboptimal demonstrations may be further compromised. 

\subsection{LfD with demonstrations collected from human teleoperation}
\begin{figure}[ht!]
    \centering
    \begin{subfigure}[t]{0.43\linewidth}
        \includegraphics[width=\linewidth]{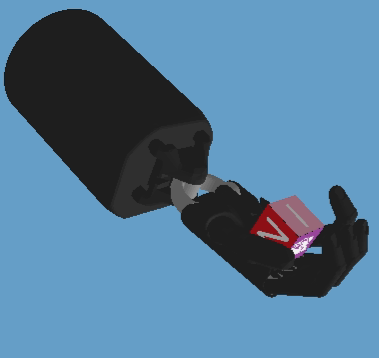}
        \caption{PyBullet shadow hand block rotation task in~\cite{zahlner2020teleoperation}.}
        \label{fig:pybullet_shadowhand_block}
    \end{subfigure}
    ~
    \begin{subfigure}[t]{0.445\linewidth}
      \includegraphics[width=\linewidth]{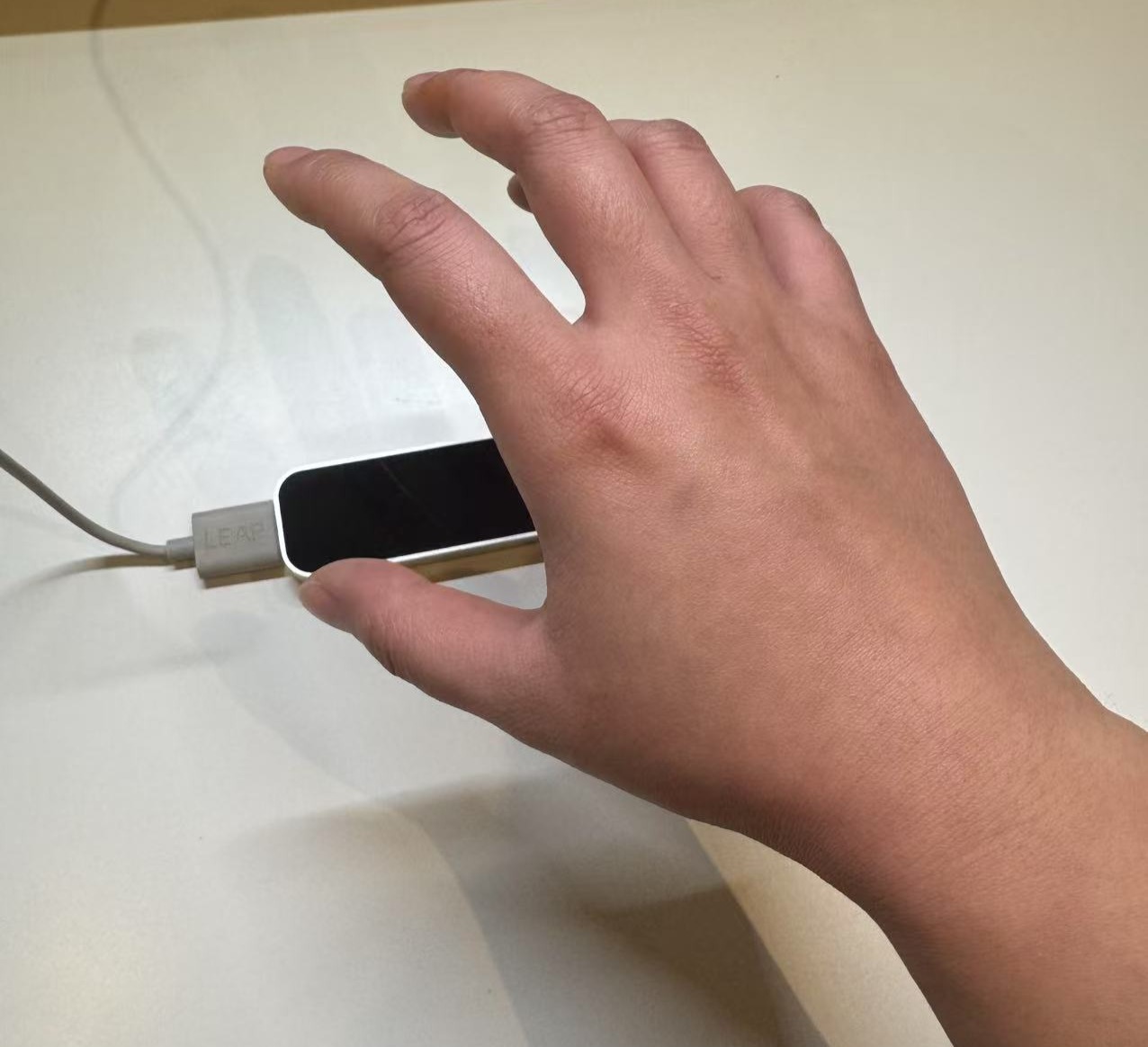}  
      \caption{Teleoperator providing a task demonstration.}
      \label{fig:Human_teleoperation_demo}
    \end{subfigure}   
    \caption{Task demonstration in a PyBullet-based simulated environment and teleoperation using a Leap Motion Controller}
\end{figure}

Next, we investigated a more challenging and realistic scenario for in-hand manipulation learning using LfD from a human expert. In this experiment, we focus on a block rotation task executed by a Shadow Hand in PyBullet, a real-time physics engine (see Figure~\ref{fig:pybullet_shadowhand_block}). Here, a block is held with a randomly assigned initial orientation, and the goal is to manipulate it until it reaches and maintains a specific target orientation within a predefined threshold. As in the Gymnasium-robotics environments, the reward signal is sparse and binary.  

\subsubsection{\textbf{Demonstration data}}
Human teleoperation is conducted using the Leap Motion Controller (LMP). As illustrated in Figure~\ref{fig:Human_teleoperation_demo}, a human teacher employs the LMP to control the Shadow Hand in the simulation and perform manipulation tasks. Real-time hand joint data is captured from the LMP at each timestep throughout an episode. 
Episodes where the original desired goal is achieved at any point during the episode and the rotation is held continuously for over five time steps are then stored in the demonstration dataset.


\begin{figure}[ht!]
    \centering
    \includegraphics[width=0.9\linewidth]{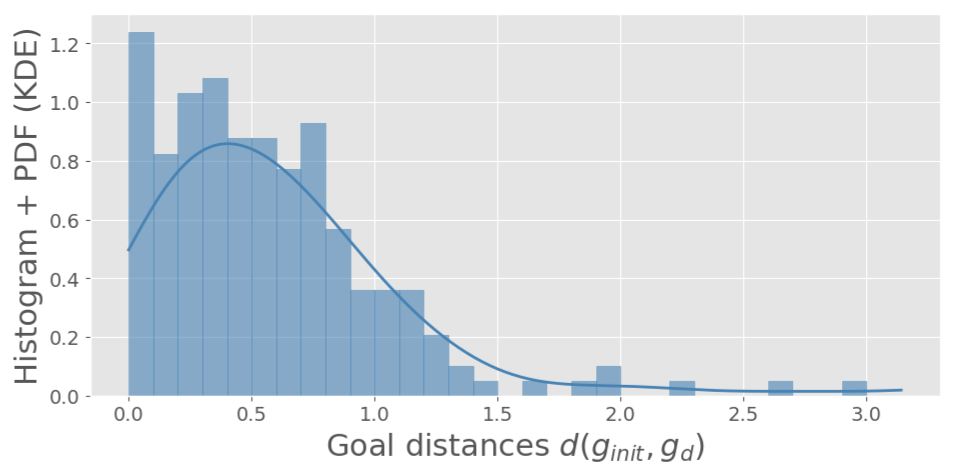}  
    \caption[The distribution of the initial and target goal distances for the human demonstration trajectories]{The distribution of the initial and target goal distances for the human demonstration trajectories: we calculate the goal distance $d(g_{init},g_d)$ between the initial pos $g_{init}$ and the target goal $g_d$ for each demonstration trajectory, and classify them into different distance intervals. The probability density of the distance intervals of all trajectories is plotted.}
    \label{fig:lfd_demo_distribution}   
\end{figure}

Before proceeding to the learning experiments, we analysed the demonstration dataset collected from human teleoperation. The teleoperation setup is relatively basic compared to more sophisticated (and more expensive) systems such as CyberGloves or VR simulations and provides limited feedback. This simplicity makes it difficult for human demonstrators to control the simulated robot hand smoothly and accurately. As a result, the average percentage of time when the final goal is achieved within an episode is only 24\% in these human demonstrations, compared to 85\% for a well-trained RL agent. This disparity indicates that the human teacher often struggles to complete the task smoothly, leading to demonstrations of relatively lower quality than those provided by an optimal RL-trained agent.

Additionally, we examined the distribution of goal distances $d(g_{init},g_d)$ between the initial positions $g_{init}$ and the original target goals $g_d$ across all trajectories. In multi-goal learning scenarios, it is crucial for LfD to benefit from demonstrations that uniformly cover the entire goal space. Ideally, the goal distances should span all ranges evenly. However, as shown in Figure~\ref{fig:lfd_demo_distribution}, the majority of human teleoperation demos involve easy, shorter goal distances, with very few examples covering challenging goal distances. This suggests that while these demonstrations offer useful examples for simpler subtasks, they are not as helpful for more challenging ones.

\subsubsection{\textbf{Hyperparameters for learning}}
200 human demonstration trajectories are collected and utilised for this LfD experiment. The hyperparameters for the RL and LfD methods are consistent with those used in the Gymnasium-robotics experiments described earlier. The key difference lies in the sampling strategy during the policy update stage for LfD methods. Instead of uniformly sampling from all experiences, we deliberately increase the proportion of demonstration data by initially drawing 50\% of the experiences from the expert buffer, with the remaining sampled from the self-collected experience buffer. This percentage is then annealed during the learning process. 

\subsubsection{\textbf{Results and discussions}}

\begin{figure}
    \centering
    \includegraphics[width=0.9\linewidth]{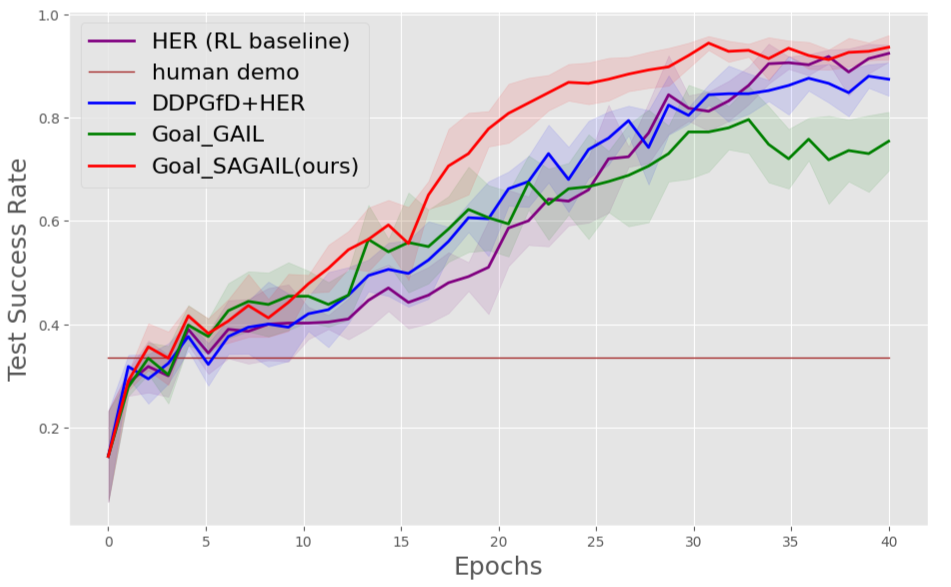}  
    \caption[The learning curve for different LfD methods with human teleop demos on PyBullet HandBlockRotate environment.]{Learning curves for different LfD methods with human teleop demos on the HandBlockRotate environment. The purple, blue, green and yellow lines depict the learning curves for DDPG+HER (the RL baseline), DDPGfD+HER with demonstration, Goal-GAIL, and Goal-SAGAIL, respectively. The brown line
    represents the average success rate of the human demonstration.}
    \label{fig:LfD_humandemo_learning_curve}
\end{figure}

The learning curves for different LfD methods compared with the RL baseline are illustrated in Figure~\ref{fig:LfD_humandemo_learning_curve}, with the results averaged across five random seeds for each method. 
The findings indicate that incorporating human teleoperation demonstrations helps both DDPGfD+HER and Goal-GAIL accelerate learning during the intermediate stages. However, both methods experience a slowdown in learning during the final stages, ultimately falling short of the RL baseline's performance. Notably, DDPGfD achieves a significantly lower final performance compared to the other approaches.

Conversely, Goal-SAGAIL significantly accelerates the learning process across all stages, consistently surpassing DDPGfD+HER and Goal-GAIL by a wide margin.  
Moreover, its final performance is markedly superior to both DDPGfD+HER and Goal-GAIL, ultimately matching the performance level of pure RL.

In our experiments, the human teleoperation demonstrations were markedly suboptimal, predominantly focusing on simpler sub-tasks beneficial only in the initial learning phases. As LfD algorithms' RL components begin to master these easier sub-tasks, they move on to tackle more challenging ones, rendering the human teleoperation demonstrations less instructive. Goal-SAGAIL, by augmenting the demonstration dataset with successful self-generated trajectories that surpass the original demonstrations, manages to outperform the other methods.

\section{CONCLUSION AND FUTURE WORK}


In this paper, we address the challenge of leveraging limited, suboptimal demonstrations to accelerate the learning process for multi-goal robot manipulation tasks.
We present the Goal-based Self-Adaptive Generative Adversarial Imitation Learning (Goal-SAGAIL) method, which combines the strengths of Goal-GAIL with a self-adaptive approach inspired by SAIL.
By integrating this self-adaptive mechanism, the learning process shifts from relying solely on pre-existing demonstrations to progressively leveraging high-quality, self-generated experiences as the agent’s performance exceeds that of the initial demonstrations.
This dynamic transition effectively addresses the challenge posed by suboptimal demonstration quality.
Experimental results show that Goal-SAGAIL outperforms the state-of-the-art multi-goal imitation learning method, Goal-GAIL, particularly when the available demonstrations are limited or suboptimal, such as those obtained via human teleoperation.

Currently, the method for comparing the difficulty of two trajectories in Goal-SAGAIL relies on a distance-based metric, which is primarily suited for robot manipulation tasks. In future work, we plan to enhance Goal-SAGAIL by generalising it to a wider range of multi-goal learning tasks. Specifically, we are exploring more universal approaches for comparing the difficulty of multi-goal trajectories beyond geometric distance.
On the experimental side, we plan to assess Goal-SAGAIL using more advanced demonstration collection methods across a range of robot manipulation tasks.



\bibliographystyle{IEEEtran}
\bibliography{references}


\end{document}